\documentclass[10pt,twocolumn,letterpaper]{article}

\usepackage{cvpr}              % To produce the CAMERA-READY version
\usepackage[dvipsnames]{xcolor}

\definecolor{cvprblue}{rgb}{0.21,0.49,0.74}
\usepackage[pagebackref,breaklinks,colorlinks,citecolor=cvprblue]{hyperref}

\usepackage[dvipsnames]{xcolor}
\usepackage{colortbl}
\usepackage{url}            % simple URL typesetting
\usepackage{booktabs}       % professional-quality tables
\usepackage{amsfonts}       % blackboard math symbols
\usepackage{nicefrac}       % compact symbols for 1/2, etc.
\usepackage{microtype}      % microtypography
\usepackage[dvipsnames]{xcolor}         % colors
\usepackage{xspace}
\usepackage{colortbl}
\usepackage{graphicx}
\usepackage{enumitem}
\usepackage{subcaption}
\usepackage{epigraph}
\usepackage{changepage}
\usepackage{multirow}
\usepackage{amssymb}
\usepackage{amsmath}
\usepackage{pifont}
\usepackage{appendix}
\usepackage{booktabs} % for table

\newcommand{\website}{https://walt-video-diffusion.github.io/}

\newcommand{\vldm}{W.A.L.T\xspace}
\newlength\savewidth\newcommand\shline{\noalign{\global\savewidth\arrayrulewidth
  \global\arrayrulewidth 1pt}\hline\noalign{\global\arrayrulewidth\savewidth}}
\newcommand{\tablestyle}[2]{\setlength{\tabcolsep}{#1}\renewcommand{\arraystretch}{#2}\centering\footnotesize}
\definecolor{baselinecolor}{HTML}{d6eaf8}
\newcommand{\baseline}[1]{\cellcolor{baselinecolor}{#1}}
\definecolor{mygray}{gray}{0.4}
\newenvironment{mytiny}{\color{mygray}\small}{}
\title{Photorealistic Video Generation with Diffusion Models}

\author{
Agrim Gupta$^{1, 2,*}$\;\;\;
Lijun Yu$^{2}$ \;\;
Kihyuk Sohn$^{2}$ \;\;
Xiuye Gu$^{2}$ \;\;
Meera Hahn$^{2}$\;\;
Li Fei-Fei$^{1}$ \\
Irfan Essa$^{2, 3}$ \;\;
Lu Jiang$^{2}$ \;\;
Jos\'e Lezama$^{2}$ \\[.5em]
$^{1}$ Stanford University\;\;
$^{2}$ Google Research\;\;
$^{3}$ Georgia Institute of Technology
}

\begin{document}
\twocolumn[{%
\renewcommand\twocolumn[1][]{#1}%
\maketitle
\begin{center}
    \centering
    \captionsetup{type=figure}
\includegraphics[width=1\textwidth]{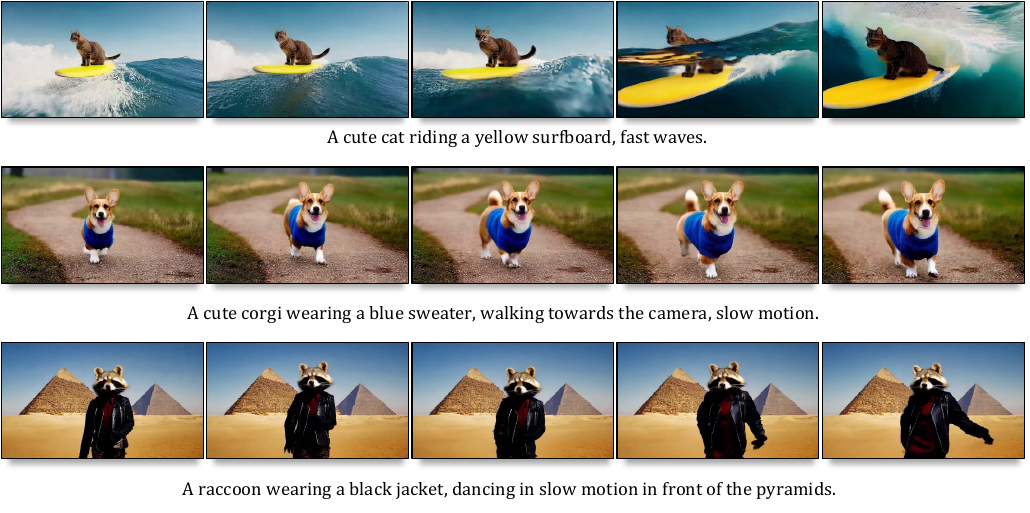}
    \captionof{figure}{\textbf{\vldm samples for text-to-video generation.} Our approach can generate high-resolution, temporally consistent photorealistic videos from text prompts. The samples shown are $512\times896$ resolution over $3.6$ seconds duration at $8$ frames per second.}
    \label{fig:teaser}
\end{center}%
}]

\begin{abstract}
We present \vldm{}, a transformer-based approach for photorealistic video generation via diffusion modeling. Our approach has two key design decisions. First, we use a causal encoder to jointly compress images and videos within a unified latent space, enabling training and generation across modalities. Second, for memory and training efficiency, we use a window attention architecture tailored for joint spatial and spatiotemporal generative modeling. Taken together these design decisions enable us to achieve state-of-the-art performance on established video (UCF-101 and Kinetics-600) and image (ImageNet) generation benchmarks without using classifier free guidance. Finally, we also train a cascade of three models for the task of text-to-video generation consisting of a base latent video diffusion model, and two video super-resolution diffusion models to generate videos of $512 \times 896$ resolution at $8$ frames per second.
\end{abstract}

\makeatletter{\renewcommand*{\@makefnmark}{}
  \footnotetext{$^*$Work partially done during an internship at Google.}\makeatother}
\section{Introduction}
\label{sec:intro}
Transformers~\cite{vaswani2017attention} are highly scalable and parallelizable neural network architectures designed to win the \textit{hardware lottery}~\cite{hooker2021hardware}. This desirable property has encouraged the research community to increasingly favor transformers over domain-specific architectures in diverse fields such as language~\cite{radford2018improving, radford2019language, radford2021learning, googlepalm2}, audio~\cite{agostinelli2023musiclm}, speech~\cite{radford2023robust}, vision~\cite{dosovitskiy2020image, he2022masked}, and robotics~\cite{brohan2022rt, brohan2023rt, bousmalis2023robocat}. Such a trend towards unification allows researchers to share and build upon advancements in traditionally disparate domains. Thus, leading to a virtuous cycle of innovation and improvement in model design favoring transformers.

A notable exception to this trend is generative modelling of videos. Diffusion models~\cite{sohl2015deep, song2019generative} have emerged as a leading paradigm for generative modelling of images~\cite{ho2020denoising, dhariwal2021diffusion} and videos~\cite{ho2022video}. However, the U-Net architecture~\cite{ronneberger2015u, ho2020denoising}, consisting of a series of convolutional~\cite{lecun1998gradient} and self-attention~\cite{vaswani2017attention} layers, has been the predominant backbone in all video diffusion approaches~\cite{ho2020denoising, dhariwal2021diffusion, ho2022video}. This preference stems from the fact that the memory demands of full attention mechanisms in transformers scale quadratically with input sequence length. Such scaling leads to prohibitively high costs when processing high-dimensional signals like video.

Latent diffusion models (LDMs)~\cite{rombach2022high} reduce computational requirements by operating in a lower-dimensional latent space derived from an autoencoder~\cite{vincent2008extracting, van2017neural, esser2021taming}. A critical design choice in this context is the type of latent space employed: spatial compression (per frame latents) versus spatiotemporal compression. Spatial compression is often preferred because it enables leveraging pre-trained image autoencoders and LDMs, which are trained on large paired image-text datasets. However, this choice increases network complexity and limits the use of transformers as backbones, especially in generating high-resolution videos due to memory constraints. On the other hand, while spatiotemporal compression can mitigate these issues, it precludes the use of paired image-text datasets, which are much larger and diverse than their video counterparts.

We present \textbf{W}indow \textbf{A}ttention \textbf{L}atent \textbf{T}ransformer (\vldm): a transformer-based method for latent video diffusion models (LVDMs). Our method consists of two stages. First, an autoencoder maps both videos and images into a unified, lower-dimensional latent space. This design choice enables training a single generative model \textit{jointly} on image and video datasets and significantly reduces the computational burden for generating high resolution videos. Subsequently, we propose a new design of transformer blocks for latent video diffusion modeling which is composed of self-attention layers that alternate between non-overlapping, window-restricted spatial and spatiotemporal attention. This design offers two primary benefits: firstly, the use of local window attention significantly lowers computational demands. Secondly, it facilitates joint training, where the spatial layers independently process images and video frames, while the spatiotemporal layers are dedicated to modeling the temporal relationships in videos.

While conceptually simple, our method provides the first empirical evidence of transformers' superior generation quality and parameter efficiency in latent video diffusion on public benchmarks. Specifically, we report state-of-the-art results on class-conditional video generation (UCF-101~\cite{soomro2012ucf101}), frame prediction (Kinetics-600~\cite{carreira2018short}) and class conditional image generation (ImageNet~\cite{deng2009imagenet}) without using classifier free guidance. Finally, to showcase the scalability and efficiency of our method we also demonstrate results on the challenging task of photorealistic text-to-video generation. We train a cascade of three models consisting of a base latent video diffusion model, and two video super-resolution diffusion models to generate videos of $512 \times 896$ resolution at $8$ frames per second and report state-of-the-art zero-shot FVD score on the UCF-101 benchmark.

%##################################################################################################
\begin{figure*}[t]
\centering
\includegraphics[width=0.9\textwidth]{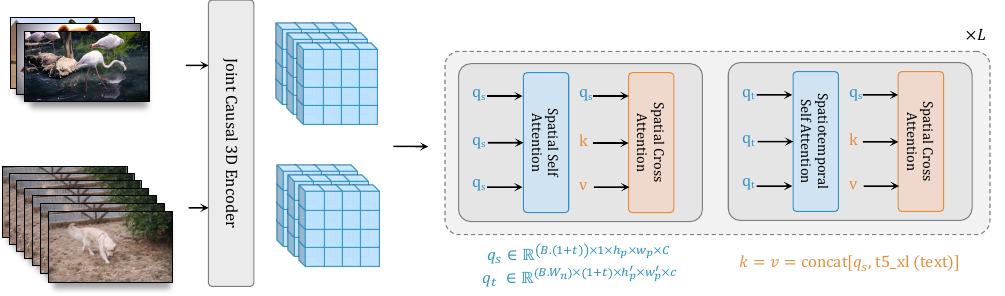}
  \caption{\textbf{\vldm.} We encode images and videos into a shared latent space. The transformer backbone processes these latents with blocks having two layers of window-restricted attention: spatial layers capture spatial relations in both images and video, while spatiotemporal layers model temporal dynamics in videos and \textit{passthrough} images via identity attention mask. Text conditioning is done via spatial cross-attention.}
\label{fig:model}
\vspace{-3mm}
\end{figure*}
%##################################################################################################
\section{Related Work}
\paragraph{Video Diffusion Models.}
Diffusion models have shown impressive results in image~\cite{sohl2015deep,song2020denoising,nichol2021improved,ho2020denoising,rombach2022high,hoogeboom2023simple} and video generation~\citep{ho2022video,ho2022imagen,singer2022make,harvey2022flexible,blattmann2023align,ge2023preserve}. Video diffusion models can be categorized into pixel-space~\citep{ho2022video,ho2022imagen,singer2022make} and latent-space~\cite{he2023latent,yu2023video,blattmann2023align,ge2023preserve} approaches, the later bringing important efficiency advantages when modeling videos. ~\citet{ho2022video} demonstrated that the quality of text conditioned video generation can be significantly improved by jointly training on image and video data. Similarly, to leverage image datasets, latent video diffusion models inflate a pre-trained image model, typically a U-Net~\cite{ronneberger2015u}, into a video model by adding temporal layers, and initializing them as the identity function \cite{ho2022imagen,singer2022make,blattmann2023align}. Although computationally efficient, this approach couples the design of video and image models, and precludes spatiotemporal compression. In this work, we operate on a unified latent space for images and videos, allowing us to leverage large scale image and video datasets while enjoying computational efficiency gains from spatiotemporal compression of videos.

\paragraph{Transformers for Generative Modeling.}
Multiple classes of generative models have utilized Transformers~\cite{vaswani2017attention} as backbone, such as, Generative adversarial networks~\cite{lee2021vitgan, jiang2021transgan, zhang2022styleswin}, autoregressive~\cite{ramesh2021zero, esser2021taming, chang2022maskgit, yu2022scaling, wu2022nuwa, gafni2022make, chang2023muse, villegas2022phenaki, yu2022magvit, gupta2022maskvit, yan2021videogpt} and diffusion~\cite{bao2022all, peebles2022scalable, gao2023masked, zheng2023fast, jabri2023scalable, VDT} models. Inspired by the success of autoregressive pretraining of large language models~\cite{radford2018improving, radford2019language, radford2021learning}, ~\citet{ramesh2021zero} trained a text-to-image generation model by predicting the next visual token obtained from an image tokenizer. Subsequently, this approach was applied to multiple applications including class-conditional image generation
~\cite{esser2021taming,yu2021vector}, text-to-image ~\cite{ramesh2021zero,yu2022scaling,ding2021cogview,wu2021godiva} or image-to-image translation~\cite{gafni2022make,wu2022nuwa}. Similarly, for video generation, transformer-based models were proposed to predict next tokens using 3D extensions of VQGAN~\cite{yan2021videogpt, ge2022long,yu2022magvit, hong2022cogvideo} or using per frame image latents~\cite{gupta2022maskvit}. Autoregressive sampling of videos is typically impractical given the very long sequences involved. To alleviate this issue, non-autoregressive sampling~\citep{chang2022maskgit,chang2023muse}, \ie parallel token prediction, has been adopted as a more efficient solution for transformer-based video generation~\citep{gupta2022maskvit, villegas2022phenaki,yu2022magvit}. Recently, the community has started adopting transformers as the denoising backbone for diffusion models in place of U-Net~\citep{hoogeboom2023simple,peebles2022scalable,zheng2023fast,chen2023pixart,VDT}. To the best of our knowledge, our work is the first successful empirical demonstration (\S~\ref{sec:exp_visual_generation}) of a transformer-based backbone for jointly training image and video latent diffusion models.

\section{Background}
\textbf{Diffusion formulation.} Diffusion models~\cite{sohl2015deep, ho2020denoising, song2019generative} are a class of generative models which learn to generate data by iteratively denoising samples drawn from a noise distribution. 
Gaussian diffusion models assume a forward noising process which gradually applies noise ($\boldsymbol{\epsilon}$) to real data ($\boldsymbol{x_0} \sim p_{\text{data}}$). Concretely, 
\begin{equation}
\boldsymbol{x_t} = \sqrt{\gamma(t)} \ \boldsymbol{x_0} + \sqrt{1 - \gamma(t)} \ \boldsymbol{\epsilon},
\end{equation}
where $\boldsymbol{\epsilon} \sim \mathcal{N}(\boldsymbol{0}, \boldsymbol{I}), t \in \left[0, 1\right]$, and $\gamma(t)$ is a monotonically decreasing function (noise schedule) from $1$ to $0$. Diffusion models are trained to learn the reverse process that inverts the forward corruptions:
\begin{equation}
\mathbb{E}_{x \sim p_{\text{data}}, t \sim \mathcal{U}(0, 1), \boldsymbol{\epsilon} \sim \mathcal{N}(\boldsymbol{0}, \boldsymbol{I})} \left[ \left\| \boldsymbol{y} - f_{\theta}(\boldsymbol{x_t}; \boldsymbol{c}, t) \right\|^2 \right],
\end{equation}
where $f_{\theta}$ is the denoiser model parameterized by a neural network, $\boldsymbol{c}$ is conditioning information e.g., class labels or text prompts, and the target $\boldsymbol{y}$ can be random noise $\boldsymbol{\epsilon}$, denoised input $\boldsymbol{x_0}$ or $\boldsymbol{v} = \sqrt{1 - \gamma(t)} \ \boldsymbol{\epsilon} - \sqrt{\gamma(t)} \ \boldsymbol{x_0}$. Following~\cite{salimans2022progressive, ho2022imagen}, we use $\boldsymbol{v}$-\textit{prediction} in all our experiments.

\textbf{Latent diffusion models (LDMs).} Processing high-resolution images and videos using raw pixels requires considerable computational resources. To address this, LDMs operate on the low dimensional latent space of a VQ-VAE~\citep{van2017neural, esser2021taming}. VQ-VAE consists of an encoder $E(x)$ that encodes an input video $x \in \mathbb{R}^{T \times H \times W \times 3}$ into a latent representation $z \in \mathbb{R}^{t \times h \times w \times c}$. The encoder downsamples the video  by a factor of $f_s = H/ h = W / w$ and $f_t = T/ t$, where $T = t = 1$ corresponds to using an image autoencoder. An important distinction from the original VQ-VAE is the absence of a codebook of quantized embeddings as diffusion models can operate on continous latent spaces. A decoder $D$ is trained to predict a reconstruction of the video, $\hat{x}$, from $z$. Following VQ-GAN~\citep{esser2021taming}, reconstruction quality can be further improved by adding adversarial~\citep{goodfellow2014generative} and perceptual losses~\citep{johnson2016perceptual, zhang2018unreasonable}. 

\section{\vldm}

\subsection{Learning Visual Tokens}
A key design decision in video generative modeling is the choice of latent space representation.
Ideally, we want a shared and unified compressed visual representation that can be used for generative modeling of both images and videos~\cite{villegas2022phenaki, yu2023language}. The unified representation is important because joint image-video learning is preferable due to a scarcity of labeled video data~\cite{ho2022imagen}, such as text-video pairs.
Concretely, given a video sequence $x \in \mathbb{R}^{(1 + T) \times H \times W \times C}$, we aim to learn a low-dimensional representation $z \in \mathbb{R}^{(1 + t) \times h \times w \times c}$ that performs spatial-temporal compression by a factor of $f_s = H/h = W/w$ in space and a factor of $f_t = T/t$ in time. To enable a unified representation for both videos and static images, the first frame is always encoded independently from the rest of the video. This allows static images $x \in \mathbb{R}^{1 \times H \times W \times C}$ to be treated as videos with a single frame, \ie $z \in \mathbb{R}^{1 \times h \times w \times c}$.

We instantiate this design with the causal 3D CNN encoder-decoder architecture of the MAGVIT-v2 tokenizer~\cite{yu2023language}. Typically the encoder-decoder consists of regular 3D convolution layers which cannot process the first frame independently~\cite{yu2022magvit, ge2022long}. This limitation stems from the fact that a regular convolutional kernel of size \( (k_t, k_h, k_w) \) will operate on \( \left\lfloor\frac{k_t-1}{2}\right\rfloor \) frames before and \( \left\lfloor\frac{k_t}{2}\right\rfloor \) frames after the input frames. \textit{Causal} 3D convolution layers solve this issue as the convolutional kernel operates on only the past \( k_t - 1 \) frames. This ensures that the output for each frame is influenced solely by the preceding frames, enabling the model to tokenize the first frame independently.

After this stage, the input to our model is a batch of latent tensors $z \in \mathbb{R}^{(1 + t) \times h \times w \times c}$ representing a single video or a stack of $1 + t$ independent images (Fig.~\ref{fig:model}). Different from~\cite{yu2023language}, our latent representation is real-valued and quantization-free. In the section below we describe how our model jointly processes a mixed batch of images and videos.

\subsection{Learning to Generate Images and Videos} \label{sec:joint_training}

\textbf{Patchify.} Following the original ViT~\cite{dosovitskiy2020image}, we ``patchify'' each latent frame independently by converting it into a sequence of non-overlapping $h_p \times w_p$ patches where $h_p = h / p$, $w_p = w / p$ and $p$ is the patch size. We use learnable positional embeddings~\cite{vaswani2017attention}, which are the sum of space and time positional embeddings. Position embeddings are added to the linear projections~\cite{dosovitskiy2020image} of the patches. Note that for images, we simply add the temporal position embedding corresponding to the first latent frame.

\noindent\textbf{Window attention.} Transformer models composed entirely of global self-attention modules incur significant compute and memory costs, especially for video tasks. For efficiency and for processing images and videos jointly we compute self-attention in windows~\cite{vaswani2017attention, gupta2022maskvit}, based on two types of non-overlapping configurations: spatial (S) and spatiotemporal (ST), \cf
~Fig.~\ref{fig:model}. \textit{Spatial Window (SW)} attention is restricted to all the tokens within a latent frame of size $1 \times h_p \times w_p$ (the first dimension is time). SW models the spatial relations in images and videos. \textit{Spatiotemporal Window (STW)} attention is restricted within a 3D window of size $(1 + t) \times h_p' \times h_w'$, modeling the temporal relationships among video latent frames. For images, we simply use an \textit{identity} attention mask ensuring that the \textit{value}
embeddings corresponding to the image frame latents are passed through the layer as is. Finally, in addition to absolute position embeddings we also use relative position embeddings~\cite{liu2021swin}.

Our design, while conceptually straightforward, achieves computational efficiency and enables joint training on image and video datasets. In contrast to methods based on frame-level autoencoders~\cite{ge2023preserve, blattmann2023align, gupta2022maskvit}, our approach does not suffer from flickering artifacts, which often result from encoding and decoding video frames independently. However, similar to~\citet{blattmann2023align}, we can also potentially leverage pre-trained image LDMs with transformer backbones by simply interleaving STW layers.

\subsection{Conditional Generation} To enable controllable video generation, in addition to conditioning on timestep $t$, diffusion models are often conditioned on additional conditional information $\boldsymbol{c}$ such as class labels, natural language, past frames or low resolution videos. In our transformer backbone, we incorporate three types of conditioning mechanisms as described in what follows:

\textbf{Cross-attention.} In addition to self-attention layers in our window transformer blocks, we add a cross-attention layer for text conditioned generation. When training models on just videos, the cross-attention layer employs the same window-restricted attention as the self-attention layer, meaning S/ST blocks will have SW/STW cross-attention layers (Fig.~\ref{fig:model}). However, for joint training, we only use SW cross-attention layers. For cross-attention we concatenate the input signal (query) with the conditioning signal (key, value) as our early experiments showed this improves performance.

\textbf{AdaLN-LoRA.} Adaptive normalization layers are an important component in a broad range of generative and visual synthesis models~\cite{nichol2021improved, dhariwal2021diffusion, perez2018film, karras2019style, dumoulin2016learned, peebles2022scalable}. A simple way to incorporate adaptive layer normalization is to include for each layer $i$, an MLP layer to regress a vector of conditioning parameters $A^i = \texttt{MLP}(\boldsymbol{c} + \boldsymbol{t})$, where $A^i = \texttt{concat}(\gamma_1, \gamma_2, \beta_1, \beta_2, \alpha_1, \alpha_2)$, $A^i \in \mathbb{R}^{6 \times d_{\texttt{model}}}$, and $\boldsymbol{c} \in \mathbb{R}^{d_{\texttt{model}}}$, $\boldsymbol{t} \in \mathbb{R}^{d_{\texttt{model}}}$ are the condition and timestep embeddings. In the transformer block, $\gamma$ and $\beta$ scale and shift the inputs of the multi-head attention and MLP layers, respectively, while $\alpha$ scales the output of both the multi-head attention and MLP layers. The parameter count of these additional MLP layers scales linearly with the number of layers and quadratically with the model's dimensional size ($\texttt{num\_blocks} \times d_{\texttt{model}} \times 6 \times d_{\texttt{model}}$). For instance, in a ViT-g model with $1$B parameters, the MLP layers contribute an additional $475$M parameters. Inspired by \cite{hu2021lora}, we propose a simple solution dubbed \textit{AdaLN-LoRA}, to reduce the model parameters. For each layer, we regress conditioning parameters as
\begin{align}
    \small A^1 &= \small \texttt{MLP}(\boldsymbol{c} + \boldsymbol{t}), &
    \small A^i &= \small A^1 + W_b^iW_a^i(\boldsymbol{c} + \boldsymbol{t}) \quad \forall i \neq 1,
\end{align}
where $W_b^i \,{\in}\, \mathbb{R}^{d_{\texttt{model}} \times r}$,  $W_a^i \,{\in}\, \mathbb{R}^{r \times (6 \times d_{\texttt{model}})}$. This reduces the number of trainable model parameters significantly when $r \,{\ll}\, d_{\texttt{model}}$. For example, a ViT-g model with $r \,{=}\, 2$ reduces the MLP parameters from $475$M to $12$M.

\textbf{Self-conditioning.} 
In addition to being conditioned on external inputs, iterative generative algorithms can also be conditioned on their own previously generated samples during inference~\cite{bengio2015scheduled, savinov2021step, chen2022analog}. Specifically, \citet{chen2022analog} modify the training process for diffusion models, such that with some probability $p_{\text{sc}}$ the model first generates a sample $\boldsymbol{\tilde{z}_0} = f_{\theta}(\boldsymbol{z_t}; \boldsymbol{0}, \boldsymbol{c}, t)$ and then refines this estimate using another forward pass conditioned on this initial sample: $f_{\theta}(\boldsymbol{z_t}; \texttt{stopgrad}(\boldsymbol{\tilde{z}_0}), \boldsymbol{c}, t)$. With probability $1 - p_{\text{sc}}$, only a single forward pass is done. We concatenate the model estimate with the input along the channel dimension and found this simple technique to work well when used in conjunction with $\boldsymbol{v}$-\textit{prediction}.

%##################################################################################################
% Video Gen Benchmarks
%##################################################################################################
\begin{table}[t]
\centering
\resizebox{\linewidth}{!}{
\centering
\begin{tabular}{@{}l@{\hspace{5pt}}l@{\hspace{5pt}}c@{\hspace{5pt}}c@{\hspace{5pt}}c@{\hspace{3pt}}c@{}}
\toprule
Method                                     & K600 FVD$\downarrow$  & UCF FVD$\downarrow$  & params.   & steps  \\ \midrule
TrIVD-GAN-FP~\cite{luc2020transformation}  & 25.7$_{\mytiny{\pm0.7}}$ & --                   & --         & 1        \\
Video Diffusion~\cite{ho2022video}         & 16.2$_{\mytiny{\pm0.3}}$ & --                   & 1.1B       & 256      \\
RIN~\cite{jabri2023scalable}               & 10.8                  & --                   & 411M       & 1000     \\
TATS~\cite{ge2022long}                     & --                    & 332$_{\mytiny{\pm18}}$  & 321M       & 1024     \\
Phenaki~\cite{villegas2022phenaki}         & 36.4$_{\mytiny{\pm0.2}}$ & --                   & 227M       & 48       \\
MAGVIT~\cite{yu2022magvit}                 & 9.9$_{\mytiny{\pm0.3}}$  & 76$_{\mytiny{\pm2}}$    & 306M       & 12       \\ 
MAGVITv2~\cite{yu2023language}             & 4.3$_{\mytiny{\pm0.1}}$  & 58$_{\mytiny{\pm2}}$    & 307M       & 24       \\ 
\midrule
\vldm-L \emph{Ours}                            & \textbf{3.3$_{\mytiny{\pm0.0}}$}  & 46$_{\mytiny{\pm2}}$   & 313M       & 50       \\ 
\vldm-XL \emph{Ours}                           & --                    & \textbf{36$_{\mytiny{\pm2}}$}   & 460M       & 50       \\ 
\end{tabular}
}
\caption{\textbf{Video generation} evaluation on frame prediction on Kinetics-600 and class-conditional generation on UCF-101.}
\label{tab:gen_k600}
\end{table}
%##################################################################################################

\subsection{Autoregressive Generation} \label{sec:method_auto_gen}
For generating long videos via autoregressive prediction we also train our model \textit{jointly} on the task of \textit{frame prediction}. This is achieved by conditioning the model on past frames with a probability of $p_{\text{fp}}$ during training. Specifically, the model is conditioned using $c_{\text{fp}} = \texttt{concat}(m_{\text{fp}} \circ \boldsymbol{z_t}, m_{\text{fp}})$, where $m_{\text{fp}}$ is a binary mask. The binary mask indicates the number of past frames used for conditioning. We condition on either $1$ latent frame (image to video generation) or $2$ latent frames (video prediction). This conditioning is integrated into the model through concatenation along the channel dimension of the noisy latent input. During inference, we use standard classifier-free guidance with $c_{\text{fp}}$ as the conditioning signal.

\subsection{Video Super Resolution}
Generating high-resolution videos with a single model is computationally prohibitive. Following \cite{ho2022cascaded}, we use a cascaded approach with three models operating at increasing resolutions. Our base model generates videos at $128 \times 128$ resolution which are subsequently upsampled twice via two super resolution stages. We first spatially upscale the low resolution input  $\boldsymbol{z^{\text{lr}}}$ (video or image) using a depth-to-space convolution operation. Note that, unlike training where ground truth low-resolution inputs are available, inference relies on latents produced by preceding stages (\cf teaching-forcing).
To reduce this discrepancy and improve the robustness of the super-resolution stages in handling artifacts generated by lower resolution stages, we use noise conditioning augmentation~\cite{ho2022cascaded}. Concretely, noise is added in accordance with $\gamma(t)$, by sampling a noise level as $t_\text{sr} \sim \mathcal{U}(0, t_\text{max\_noise})$ and is provided as input to our \textit{AdaLN-LoRA} layers.
\begin{table}[tp]
\centering
\label{tab:generation}
\resizebox{0.95\linewidth}{!}{%
\begin{tabular}{@{}l@{\hspace{10pt}}l@{\hspace{10pt}}c@{\hspace{10pt}}c@{\hspace{10pt}}c@{\hspace{10pt}}c@{}}
\toprule
Method & Cost (Iter$\times$BS) & FID$\downarrow$ & IS$\uparrow$ & params. & steps \\ 
\midrule
BigGAN-deep~\citep{brock2018large}    & -             & 6.95  & 171.4  & 160M & 1 \\
LDM-4~\citep{rombach2022high}         & 178k$\times$1200 & 10.56 & 103.5  & 400M & 250 \\
DiT-XL/2~\citep{peebles2022scalable}  & 7000k$\times$256  & 9.62  & 121.5  & 675M & 250 \\
ADM~\citep{dhariwal2021diffusion}   & -             & 7.49  & 127.5  & 608M & 2000 \\
MDT~\citep{gao2023masked}             & 6500k$\times$256  & 6.23  & 143.0  & 676M & 250 \\
MaskDiT~\citep{zheng2023fast}         & 1200k$\times$1024  & 5.69  & 178.0  & 736M & 40 \\ 
RIN~\citep{jabri2023scalable}         & 600k$\times$1024  & 3.42  & 182.0  & 410M & 1000 \\
\textcolor{gray}{simple diffusion}~\citep{hoogeboom2023simple} & \textcolor{gray}{500K$\times$2048} & \textcolor{gray}{2.77}  & \textcolor{gray}{211.8}  & \textcolor{gray}{2B}   & \textcolor{gray}{512} \\
\textcolor{gray}{VDM++}~\citep{kingma2023vdm}           & -             & \textcolor{gray}{2.40}  & \textcolor{gray}{225.3}  & \textcolor{gray}{2B}   & \textcolor{gray}{512} \\
\midrule
\vldm-L \emph{Ours}                    & 437k$\times$1024 & \textbf{2.56}  & \textbf{215.1} & 460M & 50 \\
\end{tabular}
}
\caption{\textbf{Class-conditional image generation on ImageNet 256$\times$256}. We adopt the evaluation protocol and implementation of ADM~\cite{dhariwal2021diffusion} and report results without classifier free guidance.}
\label{tab:imagenet}
\end{table}

\noindent\textbf{Aspect-ratio finetuning.} To simplify training and leverage broad data sources with different aspect ratios, we train our base stage using a square aspect ratio. We fine-tune the base stage on a subset of data to generate videos with a $9:16$ aspect ratio by interpolating position embeddings.

\section{Experiments}

In this section, we evaluate our method on multiple tasks: class-conditional image and video generation, frame prediction and text conditioned video generation and perform extensive ablation studies of different design choices. For qualitative results, see Fig.~\ref{fig:teaser}, Fig.~\ref{fig:exp_qual_samples}, Fig.~\ref{fig:long_3d_qual} and videos on our \href{\website}{project website}. See appendix for additional details.

\subsection{Visual Generation} \label{sec:exp_visual_generation}
\textbf{Video generation.}
We consider two standard video benchmarks,
UCF-101~\cite{soomro2012ucf101} for class-conditional generation and Kinetics-600~\cite{carreira2018short} for video prediction with $5$ conditioning frames. We use FVD~\citep{unterthiner2018towards}  as our primary evaluation metric. Across both datasets, \vldm \textit{significantly} outperforms all prior works (\cref{tab:gen_k600}). Compared to prior video diffusion models, we achieve state-of-the-art performance with less model parameters, and require $50$ DDIM~\cite{song2020denoising} inference steps.
%#################################################
% Main ablation table
%#################################################
\begin{table*}[t]
\vspace{-.2em}
\centering
%#################################################
% Patch Size
%#################################################
\captionsetup{font=small,labelfont=small}
\subfloat[
\textbf{Patch size}. Lower patch size is significantly better.
\label{tab:patch_size}
]{
\centering
\begin{minipage}{0.28\linewidth}{\begin{center}
\tablestyle{3pt}{1.05}
% \scriptsize
\begin{tabular}{ccc}
% \toprule
patch size $p$ & FVD$\downarrow$ & IS$\uparrow$ \\
\midrule
1 & \baseline{60.7}  & \baseline{87.2} \\
2 & 134.4 & 82.2 \\
4 & 461.8 & 63.9 \\
& & \\
\end{tabular}
\end{center}}\end{minipage}
}
\hspace{1em}
%#################################################
% Window
%#################################################
\subfloat[
\textbf{Spatiotemporal window size}. Full self-attention is not essential for good performance. sps is steps per sec.
\label{tab:window}
]{
\centering
\begin{minipage}{0.36\linewidth}{\begin{center}
\tablestyle{3pt}{1.05}
% \scriptsize
\begin{tabular}{ccccc}
% \toprule
st window & FVD$\downarrow$ & IS$\uparrow$ & sps\\
\midrule
$5\times4\times4$   & 56.9  & 87.3  & 2.24 \\
$5\times8\times8$   & \baseline{59.6}  & \baseline{87.4}  & 2.00 \\
$5\times16\times16$ & 55.3  & 87.4  & 1.75 \\\midrule
full self attn.     & 59.9  & 87.8  & 1.20 \\
& & \\
\end{tabular}
\end{center}}\end{minipage}
}
\hspace{1em}
%#################################################
% Self-cond
%#################################################
\subfloat[
\textbf{Self-conditioning}. Higher $p_{\text{sc}}$ is better.
\label{tab:self_cond}
]{
\centering
\begin{minipage}{0.28\linewidth}{\begin{center}
\tablestyle{3pt}{1.05}
% \scriptsize
\begin{tabular}{ccc}
% \toprule
$p_{\text{sc}}$ & FVD$\downarrow$ & IS$\uparrow$ \\
\midrule
0.0 & 109.9 & 82.6 \\
0.3 & 76.0 & 86.5 \\
0.6 & 60.0 & 86.8 \\
0.9 & \baseline{61.4} & \baseline{87.1} \\
& & \\
\end{tabular}
\end{center}}\end{minipage}
}
\vspace{-.1em}
%#################################################
% AdaLN Lora
%#################################################
\captionsetup{font=small,labelfont=small}
\subfloat[
\textbf{AdaLN-LoRA}. Bigger $r$ is better.
\label{tab:adaln_lora}
]{
\centering
\begin{minipage}{0.28\linewidth}{\begin{center}
\tablestyle{3pt}{1.05}
% \scriptsize
\begin{tabular}{cccc}
% \toprule
$r$ & FVD$\downarrow$ & IS$\uparrow$ & params\\
\midrule
2   & \baseline{60.7} & \baseline{87.2} & 313 M \\
4   & 56.6 & 87.3 & 314 M \\
16  & 55.5 & 88.0 & 316 M \\
64  & 54.4 & 87.9 & 324 M \\
256 & 52.5 & 88.5 & 357 M \\
\end{tabular}
\end{center}}\end{minipage}
}
\hspace{1em}
%#################################################
% Other improvements
%#################################################
\subfloat[
\textbf{Other improvements}. See text for details.
\label{tab:misc_improvements}
]{
\centering
\begin{minipage}{0.28\linewidth}{\begin{center}
\tablestyle{3pt}{1.05}
% \scriptsize
\begin{tabular}{lcc}
% \toprule
 & FVD$\downarrow$ & IS$\uparrow$ \\
\midrule
w/o qk norm~\cite{dehghani2023scaling}     & 59.0 & 86.8 \\
w/o latent norm & 67.9 & 87.1 \\
w/o zero snr~\cite{lin2023common}     & 91.0 & 84.2 \\\midrule
full method     & \baseline{60.7} & \baseline{87.2} \\
\end{tabular}
\end{center}}\end{minipage}
}
\hspace{1em}
%#################################################
% Tokenizer
%#################################################
\subfloat[
\textbf{Latent dimension $c$}. Higher $c$ is better.
\label{tab:tokenizer}
]{
\centering
\begin{minipage}{0.28\linewidth}{\begin{center}
\tablestyle{3pt}{1.05}
% \scriptsize
\begin{tabular}{cccc}
% \toprule
$c$ & rFVD$\downarrow$ & FVD$\downarrow$ & IS$\uparrow$ \\
\midrule
4  & 37.7 & 86.4 & 84.9 \\
8  & 17.1 & \baseline{75.4} & \baseline{86.3} \\
16 & 8.2  & 67.0 & 86.0 \\
32 & 3.5  & 83.4 & 82.9 \\
\end{tabular}
\end{center}}\end{minipage}
}
\vspace{-.1em}
\caption{\textbf{Ablation experiments} on UCF-101~\cite{soomro2012ucf101}. We compare FVD and inception scores to ablate important design decisions with the \colorbox{baselinecolor}{default} setting: L model, $1\times16\times16$ spatial window, $5\times8\times8$ saptiotemporal (st) window, $p_{\text{sc}} = 0.9$, $c = 8$ and $r = 2$.}
\label{tab:ablations} \vspace{-.5em}
\end{table*}
%#################################################

\noindent\textbf{Image generation.}
To verify the modeling capabilities of \vldm on the image domain, we train a version of \vldm for the standard ImageNet class-conditional setting. For evaluation, we follow ADM~\cite{dhariwal2021diffusion} and report the FID~\cite{heusel2017gans} and Inception~\cite{salimans2016improved} scores calculated on $50$K samples generated in $50$ DDIM steps. We compare (Table~\ref{tab:imagenet}) \vldm with state-of-the-art image generation methods for $256\times256$ resolution. Our model outperforms prior works without requiring specialized schedules, convolution inductive bias, improved diffusion losses, and classifier free guidance. Although VDM++~\cite{kingma2023vdm} has slightly better FID score, the model has significantly more parameters (2B). 

\subsection{Ablation Studies}\label{subsection:ablations}
We ablate \vldm to understand the contribution of various design decisions with the default settings: model L, patch size 1, $1 \times 16 \times 16$ spatial window, $5 \times 8 \times 8$ spatiotemporal window, $p_{\text{sc}} = 0.9$, $c = 8$ and $r = 2$.

\noindent\textbf{Patch size.} In various computer vision tasks utilizing ViT\cite{dosovitskiy2020image}-based models, a smaller patch size $p$ has been shown to consistently enhance performance~\cite{dosovitskiy2020image, zhai2022scaling, caron2021emerging, gupta2023siamese}. Similarly, our findings also indicate that a reduced patch size improves performance (Table~\ref{tab:patch_size}).

\noindent\textbf{Window attention.} We compare three different STW window configurations with full self-attention (Table~\ref{tab:window}). We find that local self-attention can achieve competitive (or better) performance while being significantly faster (up to $2\times$) and requiring less accelerator memory.

\noindent\textbf{Self-conditioning.} In Table~\ref{tab:self_cond} we study the influence of varying the self-conditioning rate $p_{\text{sc}}$ on generation quality. We notice a clear trend: increasing the self conditioning rate from $0.0$ (no self-conditioning) to $0.9$ improves the FVD score substantially ($44\%$).

\noindent\textbf{AdaLN-LoRA.} An important design decision in diffusion models is the conditioning mechanism. We investigate the effect of increasing the bottleneck dimension $r$ in our proposed \textit{AdaLN-LoRA} layers (Table~\ref{tab:adaln_lora}). This hyperparameter provides a flexible way to trade off between number of model parameters and generation performance. As shown in Table~\ref{tab:adaln_lora}, increasing $r$ improves performance but also increases model parameters. This highlights an important model design question: given a fixed parameter budget, how should we allocate parameters - either by using \textit{separate} AdaLN layers, or by increasing base model parameters while using \textit{shared} AdaLN-LoRA layers? We explore this in Table~\ref{tab:param_lora} by comparing two model configurations: \vldm-L with separate AdaLN layers and \vldm-XL with AdaLN-LoRA and $r = 2$. While both configurations yield similar FVD and Inception scores, \vldm-XL achieves a lower final loss value, suggesting the advantage of allocating more parameters to the base model and choosing an appropriate $r$ value within accelerator memory limits.

%##################################################################################################
% Lora param matched ablation
%##################################################################################################
\begin{table}[t]
\centering
\tablestyle{5pt}{1.3}
\begin{tabular}{clcccc}
\toprule
Model & AdaLN & FVD$\downarrow$ & IS$\uparrow$ & params. & final loss\\
\shline
% L  & lora-2        & 46.2 & 89.1 & 313M  & 0.275 \\
L  & separate      & 34.6 & 90.2 & 458M  & 0.274 \\
XL & LoRA-2        & 36.7 & 89.4 & 460M  & 0.268 \\
\end{tabular}
\caption{\label{tab:param_lora}\textbf{Parameter matched} comparison between AdaLN-LoRA and per layer adaln layers. See text for details.}
\vspace{-2em}
\end{table}
%##################################################################################################

%##################################################################################################
\begin{figure*}[t]
\centering
\includegraphics[width=1\textwidth]{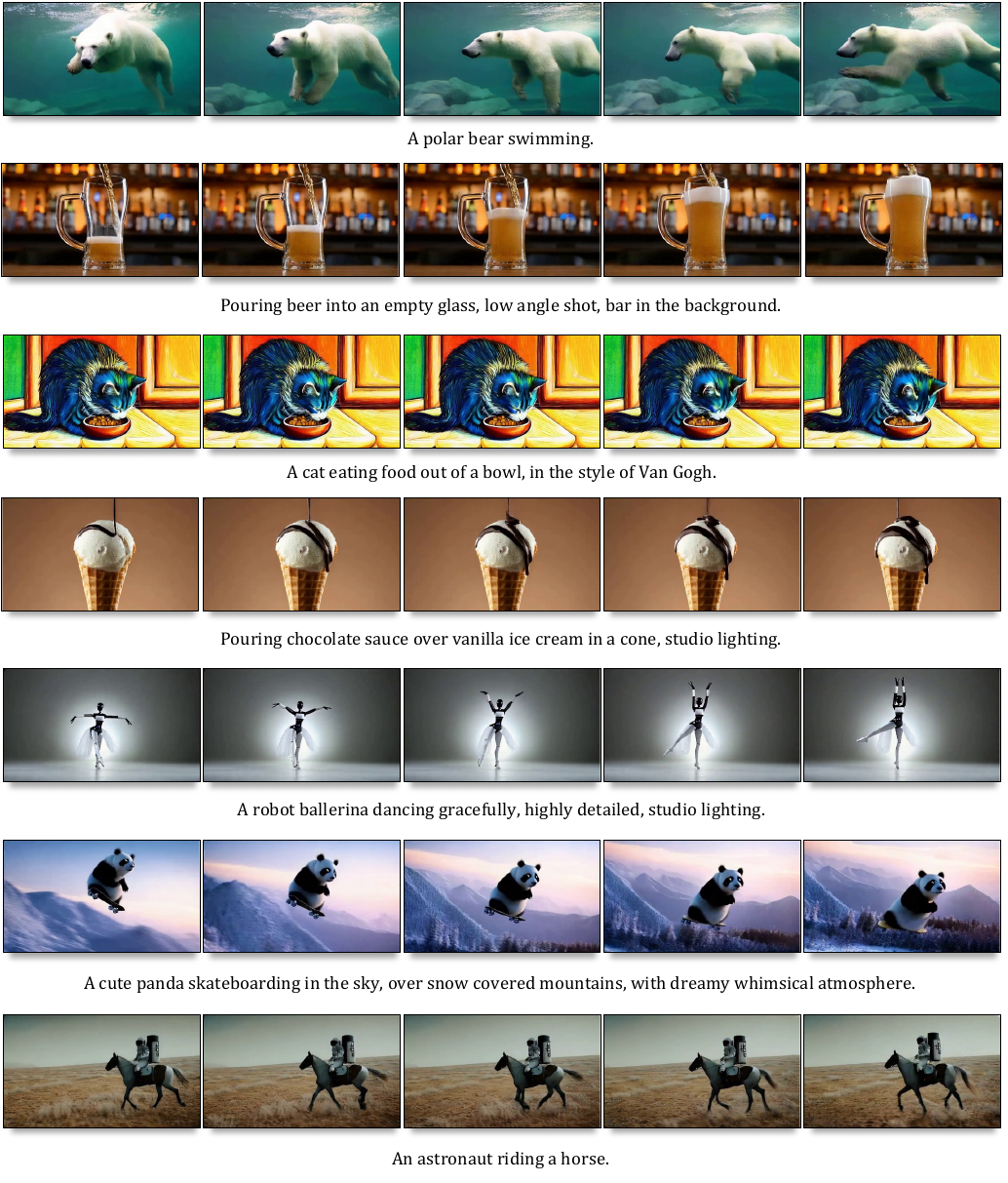}
  \caption{\textbf{Qualitative evaluation.} Example videos generated by \vldm from natural language prompts at $512\times896$ resolution over $3.6$ seconds duration at $8$ frames per second. The \vldm model is able to generate temporally consistent photorealistic videos that align with the textual prompt.}
\label{fig:exp_qual_samples}
\vspace{-3mm}
\end{figure*}
%##################################################################################################

\noindent\textbf{Noise schedule.} Common latent diffusion noise schedules~\cite{rombach2022high} typically do not ensure a zero signal-to-noise ratio (SNR) at the final timestep, i.e., at $t = 1, \gamma(t) > 0$. This leads to a mismatch between training and inference phases. During inference, models are expected to start from purely Gaussian noise, whereas during training, at $t = 1$, a small amount of signal information remains accessible to the model. This is especially harmful for video generation as videos have high temporal redundancy. Even minimal information leakage at $t = 1$ can reveal substantial information to the model. Addressing this mismatch by enforcing a zero terminal SNR~\cite{lin2023common} significantly improves performance (Table~\ref{tab:misc_improvements}). Note that this approach was originally proposed to fix over-exposure problems in image generation, but we find it effective for video generation as well.

\noindent\textbf{Autoencoder.} Finally, we investigate one critical but often overlooked hyperparameter in the first stage of our model: the channel dimension $c$ of the autoencoder latent $z$. As shown in Table~\ref{tab:tokenizer}, increasing $c$ significantly improves the reconstruction quality (lower rFVD) while keeping the same spatial $f_s$ and temporal compression $f_t$ ratios.  Empirically, we found that both lower and higher values of $c$ lead to poor FVD scores in generation, with a sweet spot of $c = 8$ working well across most datasets and tasks we evaluated. We also normalize the latents before processing them via transformer which further improves performance. 

In our transformer models, we use query-key normalization~\cite{dehghani2023scaling} as it helps stabilize training for larger models. Finally, we note that some of our default settings are not optimal, as indicated by ablation studies. These defaults were chosen early on for their robustness across datasets, though further tuning may improve performance.

%##################################################################################################
\begin{figure}[t]
\centering
\includegraphics[width=0.39\textwidth]{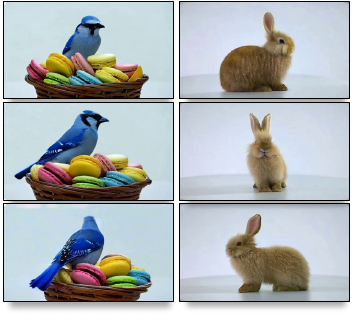}
  \caption{\textbf{Examples of consistent 3D camera motion (5.1 secs).} Prompts: \emph{camera turns around a \{blue jay, bunny\}, studio lighting, $360^{\circ}$ rotation}. Best viewed in video format. }
\label{fig:long_3d_qual}
\vspace{-3mm}
\end{figure}
%##################################################################################################

\subsection{Text-to-video}

We train \vldm for text-to-video jointly on text-image and text-video pairs (\cref{sec:joint_training}). We used a dataset of $\sim$970M text-image pairs and $\sim$89M text-video pairs from the public internet and internal sources. We train our base model at resolution $17\times128\times128$ (3B parameters), and two $2\times$ cascaded super-resolution models for $17\times128\times224 \rightarrow 17\times256\times448$ (L, 1.3B, $p=2$) and $17\times256\times448\rightarrow17\times512\times896$ (L, 419M, $p=2$) respectively. We fine-tune the base stage for the $9:16$ aspect ratio to generate videos at resolution $128\times224$. We use classifier free guidance for all our text-to-video results.

\subsubsection{Quantitative Evaluation}

Evaluating text-conditioned video generation systems scientifically remains a significant challenge, in part due to the absence of standardized training datasets and benchmarks. So far we have focused our experiments and analyses on the standard academic benchmarks, which use the same training data to ensure controlled and fair comparisons. Nevertheless, to compare with prior work on text-to-video, we also report results on the UCF-101 dataset in the zero-shot evaluation protocol in Table~\ref{tab:ucf}~\cite{ge2023preserve, singer2022make, hong2022cogvideo}. Also see supplement.

\textbf{Joint training.} A primary strength of our framework is its ability to train simultaneously on both image and video datasets. In Table~\ref{tab:ucf} we ablate the impact of this joint training approach. Specifically, we trained two versions of \vldm-L (each with $419$M params.) models using the default settings specified in \S~\ref{subsection:ablations}. We find that joint training leads to a notable improvement across both metrics. Our results align with the findings of~\citet{ho2022video}, who demonstrated the benefits of joint training for pixel-based video diffusion models with U-Net backbones.

\textbf{Scaling.} Transformers are known for their ability to scale effectively~ in many tasks~\cite{radford2018improving, dehghani2023scaling, bousmalis2023robocat}. In Table~\ref{tab:ucf} we show the benefits of scaling our transformer model for video diffusion. Scaling our base model size leads to significant improvements on both the metrics. It is important to note, however, that our base model is considerably smaller than leading text-to-video systems. For instance,~\citet{ho2022imagen} trained base model of $5.7$B parameters. Hence, we believe scaling our models further is an important direction of future work.
%##################################################################################################
% T2V UCF
%##################################################################################################
\begin{table}[t]
    \centering
    \resizebox{1\linewidth}{!}{%
        \begin{tabular}{l c c}
            \toprule
            \textbf{Method}  & IS ($\uparrow$) & FVD ($\downarrow$)  \\
            \midrule
            % CogVideo (Chinese)~\cite{hong2022cogvideo} & 23.55 & 751.34 \\
            % CogVideo (English)~\cite{hong2022cogvideo} & 25.27 & 701.59 \\
            % MagicVideo~\cite{zhou2022magicvideo}       & - & 699.00 \\
            % Make-A-Video~\cite{singer2022make}         & 33.00 & 367.23 \\
            % Video LDM~\cite{blattmann2023align}        & 33.45 & 550.61 \\
            % PYoCo~\cite{ge2023preserve}                & 47.76 & 355.19 \\
            CogVideo (Chinese)~\cite{hong2022cogvideo} & 23.6 & 751.3 \\
            CogVideo (English)~\cite{hong2022cogvideo} & 25.3 & 701.6 \\
            MagicVideo~\cite{zhou2022magicvideo}       & - & 699.0 \\
            Make-A-Video~\cite{singer2022make}         & 33.0 & 367.2 \\
            Video LDM~\cite{blattmann2023align}        & 33.5 & 550.6 \\
            PYoCo~\cite{ge2023preserve}                & \textbf{47.8} & 355.2 \\
            \midrule
            \vldm \emph{(Ours)} 419M (video only)      & 26.8 & 598.8 \\
            \vldm \emph{(Ours)} 419M (video + image)   & 31.7 & 344.5 \\
            \vldm \emph{(Ours)} 3B (video + image)     & 35.1 & \textbf{258.1} \\
        \end{tabular}
    }
    \caption{\small \textbf{UCF-101 text-to-video generation}. Joint training on image and video datasets in conjunction with scaling the model parameters is essential for high quality video generation.}
    \label{tab:ucf}
\end{table}

%##################################################################################################
\textbf{Comparison with prior work.} In Table~\ref{tab:ucf}, we present a system-level comparison of various text-to-video generation methods. Our results are promising; we surpass all previous work in the FVD metric. In terms of the IS, our performance is competitive, outperforming all but PYoCo~\cite{ge2023preserve}. A possible explanation for this discrepancy might be PYoCo's use of stronger text embeddings. Specifically, they utilize both CLIP~\cite{radford2021learning} and T5-XXL~\cite{roberts2019exploring} encoders, whereas we employ a T5-XL~\cite{roberts2019exploring} text encoder only.

\subsubsection{Qualitative Results}
As mentioned in \S~\ref{sec:method_auto_gen}, we jointly train our model on the task of frame prediction conditioned on $1$ or $2$ latent frames. Hence, our model can be used for animating images (\textbf{image-to-video}) and generating longer videos with consistent camera motion (Fig.~\ref{fig:long_3d_qual}). See videos on our \href{\website}{project website}.

\section{Conclusion}
In this work, we introduce \vldm, a simple, scalable, and efficient transformer-based framework for latent video diffusion models. We demonstrate state-of-the-art results for image and video generation using a transformer backbone with windowed attention. We also train a cascade of three \vldm models jointly on image and video datasets, to synthesize high-resolution, temporally consistent photorealistic videos from natural language descriptions. While generative modeling has seen tremendous recent advances for images, progress on video generation has lagged behind. We hope that scaling our unified framework for image and video generation will help close this gap.

\section*{Acknowledgements}
We thank 
Bryan Seybold,
Dan Kondratyuk,
David Ross,
Hartwig Adam,
Huisheng Wang,
Jason Baldridge,
Mauricio Delbracio
and Orly Liba
for helpful discussions and feedback.

%\clearpage
{
    \small
    \bibliographystyle{ieeenat_fullname}
    \bibliography{main}
}

\clearpage
\appendix
\setcounter{page}{1}
% \maketitlesupplementary

% it is now fixed
\begin{table}[t]
\centering
\resizebox{0.50\linewidth}{!}{
\centering
\begin{tabular}{@{}l@{\hspace{15pt}}r@{}}
\toprule
                                      & T2V (base)               \\
\midrule
Input                                 & $5\times16\times 28$     \\
Spatial window                        & $1\times16\times 28$     \\
Spatiotermporal window                & $5\times8\times 14$      \\
Training steps                        & 250000                   \\
Batch size                            & 512                      \\
lr schedule                           & Constant                 \\
Optimizer                             & Adafactor                \\
lr                                    & 0.00005                  \\
\end{tabular}
}
\caption{\textbf{Hyperparameters for aspect-ratio finetuning.}}
\label{tab:as-ft-hyperparm}
\end{table}

\section{Implementation Details}

For the first stage, we follow the architecture and hyperparameters from \citet{yu2023language}. We report hyperparameters specific for training our model in Table~\ref{tab:hyperparams}. To train the second stage transformer model, we use the default settings of $1 \times 16 \times 16$ spatial window, $5 \times 8 \times 8$ spatiotemporal window, $p_{\text{sc}} = 0.9$, $c = 8$ and $r = 2$. We summarize additional training and inference hyperparameters for all tasks in  Table~\ref{tab:hyperparams}. The UCF-101 model results reported in Tables~\ref{tab:gen_k600} and \ref{tab:param_lora} are trained for $60,000$ steps. We perform all ablations on UCF-101 with $35,000$ training steps.

\textbf{Aspect-ratio finetuning.} 
To simplify training and leverage broad data sources with different aspect ratios, we train the base stage using a square aspect ratio. We fine-tune the base the stage on a subset of data to generate videos with a $9:16$ aspect ratio. We interpolate the absolute and relative position embeddings and scale the window sizes. We summarize the finetuning hyperparameters in Table~\ref{tab:as-ft-hyperparm}.

\textbf{Long video generation.} As described in \S~\ref{sec:method_auto_gen}, we train our model jointly on the task of frame prediction. During inference, we generate videos as follows: Given a natural language description of a video, we first generate the initial $17$ frames using our base model. Next, we encode the last $5$ frames into $2$ latent frames using our causal 3D encoder. Providing $2$ latent frames as input for subsequent autoregressive generation helps ensure that our model can maintain continuity of motion and produce temporally consistent videos. 

\textbf{UCF-101 Text-to-Video.} We follow the evaluation protocol of prior work~\cite{ge2023preserve}, and adapt their prompts to better describe the UCF-101 classes.

\section{Additional Results}
\subsection{Image Generation}
\begin{figure}[tp]
% \vspace{-4mm}
    \centering
    \includegraphics[width=\linewidth,trim={0 0 0 0},clip]{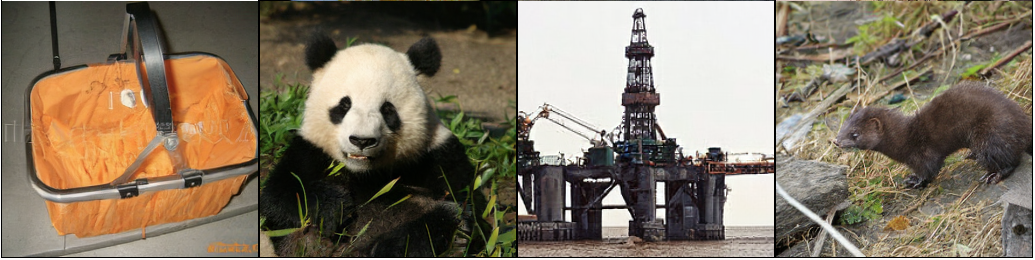}\\[-.2em]
    \includegraphics[width=\linewidth,trim={0 0 0 0},clip]{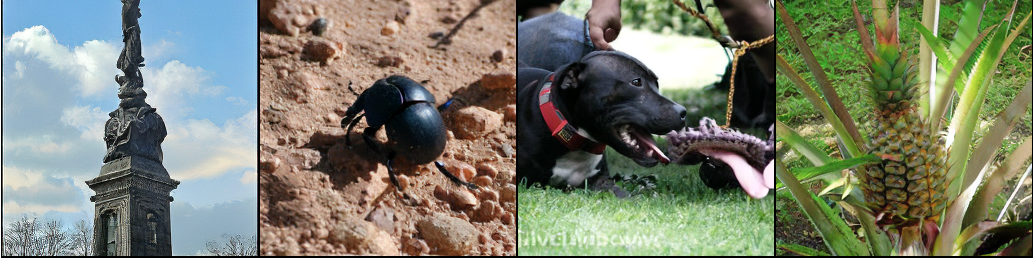}\\   
        
    \caption{\textbf{ImageNet class-conditional generation samples}. }
    \label{fig:supp_inet}
    \vspace{-2mm}
% xid/75915442#2  
\end{figure}

We compare (Table~\ref{tab:im_cfg}) \vldm with state-of-the-art image generation methods for $256\times256$ resolution with classifier free guidance. Unlike, prior work~\cite{peebles2022scalable, gao2023masked, zheng2023fast} using Transformer for diffusion modelling, we did not observe significant benefit of using vanilla classifier free guidance. Hence, we report results using the power cosine schedule proposed by ~\citet{gao2023masked}. Our model performs better than prior works on the Inception Score metric, and achieves competitive FID scores.
\cref{fig:supp_inet} shows qualitative samples.

\begin{table}[tp]
\centering
\resizebox{\linewidth}{!}{
\begin{tabular}{@{}l@{\hspace{10pt}}r@{\hspace{10pt}}c@{\hspace{10pt}}c@{\hspace{10pt}}r@{\hspace{10pt}}r@{}}
\toprule
Method                                       & Cost (Iter$\times$BS)    & FID$\downarrow$ & IS$\uparrow$ & Params. & Steps \\
\midrule
LDM-4~\citep{rombach2022high}                & 178k$\times$1200         & 3.60            & 247.7        & 400M    & 250   \\
DiT-XL/2~\citep{peebles2022scalable}         & 7000k$\times$256         & 2.27            & 278.2        & 675M    & 250   \\
ADM~\citep{dhariwal2021diffusion}            & -                        & 3.94            & 215.8        & 608M    & 2000  \\
MDT~\citep{gao2023masked}                    & 6500k$\times$256         & 1.79            & 283.0        & 676M    & 250   \\
MaskDiT~\citep{zheng2023fast}                & 1200k$\times$1024        & 2.28            & 276.6        & 736M    & 40    \\ 
simple diffusion~\citep{hoogeboom2023simple} & 500K$\times$2048         & 2.44            & 256.3        & 2B      & 512   \\
VDM++~\citep{kingma2023vdm}                  & -                        & 2.12            & 267.7        & 2B      & 512   \\
\midrule
\vldm-L \emph{Ours}                          & 437k$\times$1024         & 2.40             & 290.5          & 460M    & 50    \\
\bottomrule
\end{tabular}
}
\caption{\textbf{Class-conditional image generation on ImageNet 256$\times$256}. We adopt the evaluation protocol and implementation of ADM~\cite{dhariwal2021diffusion} and report results \textit{with} classifier free guidance.}
\label{tab:im_cfg}
\vspace{8mm}
\end{table}

\begin{table*}[t]
\centering
\resizebox{0.9\linewidth}{!}{
\centering
\begin{tabular}{@{}l@{\hspace{8pt}}c@{\hspace{8pt}}c@{\hspace{8pt}}c@{\hspace{8pt}}c@{\hspace{8pt}}c@{\hspace{8pt}}c@{\hspace{8pt}}c@{}}
\toprule
                                      & ImageNet                  & UCF-101                  & K600                    & T2V (base)               & T2V (2$\times$)  & T2V (2$\times$2$\times$)               \\\midrule
\textit{First Stage}                  &                           &                          &                         &                          &                              &                          \\
Input                                 & $1\times256\times 256$    & $17\times128\times 128$  & $17\times128\times 128$ & $17\times128\times 128$  & $17\times256\times 448$      & $17\times512\times 896$  \\
% \multicolumn{7}{c}{}                                                                   \\\midrule
$f_s, f_t$                            & 8, -                      & 8, 4                     & 8, 4                    & 8, 4                     & 8, 4                         & 8, 4                     \\
Channels                              & 128                       & 128                      & 128                     & 128                      & 128                          & 128                      \\
Channel multiplier                    & 1,1,2,4                   & 1, 2, 2, 4               & 1, 2, 2, 4              & 1, 2, 2, 4               & 1, 2, 2, 4                   & 1, 2, 2, 4               \\
Training duration                     & 270 epochs                & 2000 epochs              & 270000 steps            & 1000000 steps            & 1000000 steps                & 1000000 steps            \\
Batch size                            & 256                       & 256                      & 256                     & 256                      & 256                          & 256                      \\
lr schedule                           & Cosine                    & Cosine                   & Cosine                  & Cosine                   & Cosine                       & Cosine                   \\
Optimizer                             & Adam                      & Adam                     & Adam                    & Adam                     & Adam                         & Adam                     \\
\toprule
\textit{Second Stage}                 &                           &                          &                         &                          &                              &                          \\
Input                                 & $1\times32\times 32$      & $5\times16\times 16$     & $5\times16\times 16$    & $5\times16\times 16$     & $5\times32\times 56$         & $5\times64\times 112$     \\
Layers                                & 24                        & 28                       & 24                      & 52                       & 40                           & 24                       \\
Hidden size                           & 1024                      & 1152                     & 1024                    & 9216                     & 1408                         & 1024                     \\
Heads                                 & 16                        & 16                       & 16                      & 16                       & 16                           & 16                       \\
Training duration                     & 350 epochs                & 60000 steps              & 360 epochs              & 550000 steps             & 675000 steps                 & 275000 steps             \\
Batch size                            & 1024                      & 256                      & 512                     & 512                      & 512                          & 512                      \\
lr schedule                           & Cosine                    & Cosine                   & Cosine                  & Cosine                   & Cosine                       & Cosine                   \\
Optimizer                             & AdamW                     & AdamW                    & AdamW                   & Adafactor                & Adafactor                    & Adafactor                \\
lr                                    & 0.0005                    & 0.0005                   & 0.0005                  & 0.0002                   & 0.0005                       & 0.0005                   \\
EMA                                   & $\checkmark$              & $\checkmark$             & $\checkmark$            & \ding{55}                & \ding{55}                    & \ding{55}                \\
Patch size                            & 1                         & 1                        & 1                       & 1                        & 2                            & 4                        \\
AdaLN-LoRA                            & \ding{55}                 & 2                        & 2                       & 2                        & 2                            & 2                        \\
\toprule
\textit{Diffusion}                    &                           &                          &                         &                          &                              &                          \\
Diffusion Steps                       & 1000                      & 1000                     & 1000                    & 1000                     & 1000                         & 1000                     \\
Noise schedule                        & Linear                    & Linear                   & Linear                  & Linear                   & Linear                       & Linear                   \\
$\beta_0$                             & 0.0001                    & 0.0001                   & 0.0001                  & 0.0001                   & 0.0001                       & 0.0001                   \\
$\beta_{1000}$                        & 0.02                      & 0.02                     & 0.02                    & 0.02                     & 0.02                         & 0.02                     \\
Sampler                               & DDIM                      & DDIM                     & DDIM                    & DDIM                     & DDIM                         & DDIM                     \\
Sampling steps                        & 50                        & 50                       & 50                      & 50                       & 50                           & 50                       \\
Guidance                              & \ding{55}                 & \ding{55}                & \ding{55}               & $\checkmark$             & $\checkmark$                 & $\checkmark$             \\
\end{tabular}
}
\caption{\textbf{Training and evaluation hyperparameters.}}
\label{tab:hyperparams}
\end{table*}

\subsection{Video Generation}
\begin{figure*}[tp]
% \vspace{-4mm}
    \centering
    \includegraphics[width=\linewidth,trim={0 0 0 0},clip]{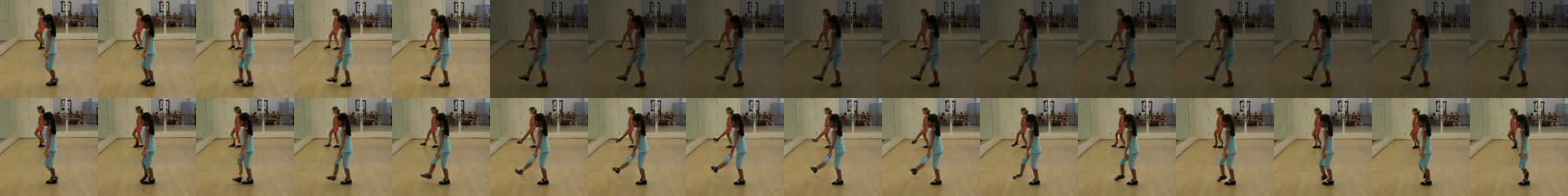}\\
    \includegraphics[width=\linewidth,trim={0 0 0 0},clip]{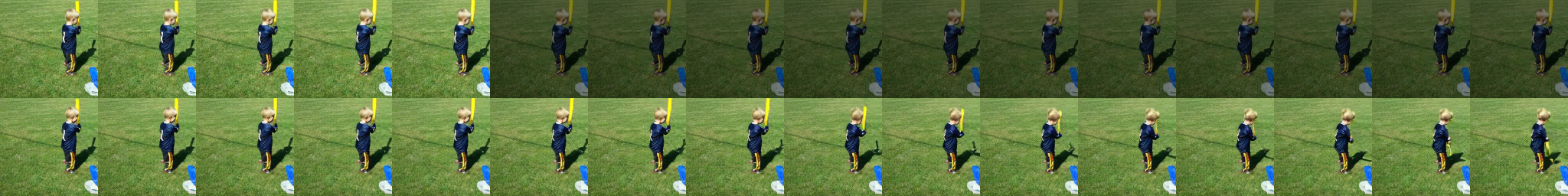}\\   
        \includegraphics[width=\linewidth,trim={0 0 0 0},clip]{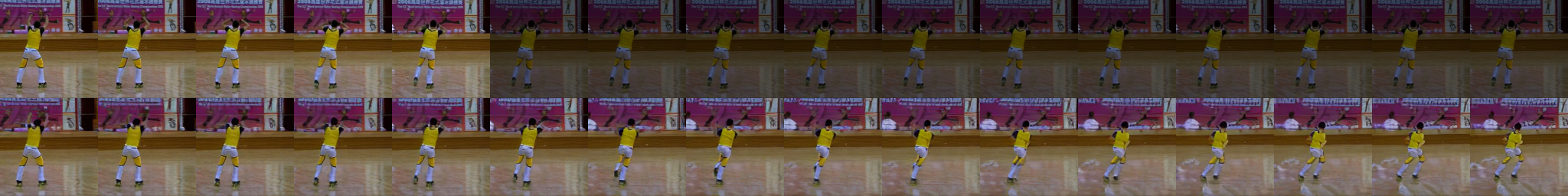}\\
        
    \caption{\textbf{Frame prediction samples on Kinetics-600}. Top: ground-truth, where unobserved frames are shaded. Bottom: generation.}
    \label{fig:supp_k600}
    % \vspace{-2mm}
% https://colab.corp.google.com/drive/1OfxCjQAYzvieeSfZGHT_yLRJqz06W-GM#scrollTo=NTigJxMmRuC-    
\end{figure*}

We show samples for Kinetics-600 frame prediction in \cref{fig:supp_k600}.

\subsection{Image-to-Video}
\vspace{4mm}
As noted in Section \ref{sec:method_auto_gen}, we train our model jointly on the task of frame prediction, where we condition on $1$ latent frame. This allows us to leverage the high quality first frame from the image generator as context for predicting subsequent frames. For qualitative results see videos on our \href{\website}{project website}.

% 
% To split the supplementary pages from the main paper, you can use \href{https://support.apple.com/en-ca/guide/preview/prvw11793/mac#:~:text=Delete%20a%20page%20from%20a,or%20choose%20Edit%20%3E%20Delete).}{Preview (on macOS)}, \href{https://www.adobe.com/acrobat/how-to/delete-pages-from-pdf.html#:~:text=Choose%20%E2%80%9CTools%E2%80%9D%20%3E%20%E2%80%9COrganize,or%20pages%20from%20the%20file.}{Adobe Acrobat} (on all OSs), as well as \href{https://superuser.com/questions/517986/is-it-possible-to-delete-some-pages-of-a-pdf-document}{command line tools}.

\end{document}